\documentclass[conference]{IEEEtran}
\IEEEoverridecommandlockouts

\usepackage{cite}
\usepackage{amsmath,amssymb,amsfonts}
\usepackage{algorithmic}
\usepackage{graphicx}
\usepackage{textcomp}
\usepackage{xcolor}

\usepackage{ulem}
\usepackage{booktabs}
\usepackage{multirow}
\usepackage{makecell}
\def\BibTeX{{\rm B\kern-.05em{\sc i\kern-.025em b}\kern-.08em
    T\kern-.1667em\lower.7ex\hbox{E}\kern-.125emX}}
\begin{document}

\title{AS-FIBA: Adaptive Selective Frequency-Injection for Backdoor Attack on Deep Face Restoration
\thanks{This work was supported in part by the National Natural Science Foundation of China (No. 62302220).
}
\thanks{* means equal contribution.}
\thanks{Jianfeng Lu (lujf@njust.edu.cn) is the corresponding author.}
}

\author{
\IEEEauthorblockN{1\textsuperscript{st} Zhenbo Song\textsuperscript{*}}
\IEEEauthorblockA{\textit{School of Computer Science and Engineering} \\
\textit{Nanjing University of Science and Technology}\\
Nanjing, China \\
songzb@njust.edu.cn}
\and
\IEEEauthorblockN{1\textsuperscript{st} Wenhao Gao\textsuperscript{*}}
\IEEEauthorblockA{\textit{School of Computer Science and Engineering} \\
\textit{Nanjing University of Science and Technology}\\
Nanjing, China \\
gaowenhao@njust.edu.cn}
\and
\IEEEauthorblockN{2\textsuperscript{nd} Zhenyuan Zhang}
\IEEEauthorblockA{\textit{School of Computer Science and Engineering} \\
\textit{Nanjing University of Science and Technology}\\
Nanjing, China \\
zyzhang.bbetter@gmail.com}
\and
\IEEEauthorblockN{3\textsuperscript{th} Jianfeng Lu}
\IEEEauthorblockA{\textit{School of Computer Science and Engineering} \\
\textit{Nanjing University of Science and Technology}\\
Nanjing, China \\
lujf@njust.edu.cn}

}

\maketitle

\begin{abstract}
Deep learning-based face restoration models, increasingly prevalent in smart devices, have become targets for sophisticated backdoor attacks. Through subtle trigger injection into input face images, these attacks can lead to unexpected restoration outcomes. Unlike conventional methods focused on classification tasks, our approach introduces a unique degradation objective tailored for attacking restoration models. Moreover, we propose the Adaptive Selective Frequency Injection Backdoor Attack (AS-FIBA) framework, employing a neural network for input-specific trigger generation in the frequency domain, seamlessly blending triggers with benign images. This results in imperceptible yet effective attacks, guiding restoration predictions towards subtly degraded outputs rather than conspicuous targets. Extensive experiments demonstrate the efficacy of the degradation objective on state-of-the-art face restoration models. Additionally,  it is notable that AS-FIBA can insert effective backdoors that are more imperceptible than existing backdoor attack methods, including WaNet, ISSBA, and FIBA.
\end{abstract}


\section{Introduction}
\begin{figure}[t]
	\centering
	\includegraphics[width=0.9\linewidth]{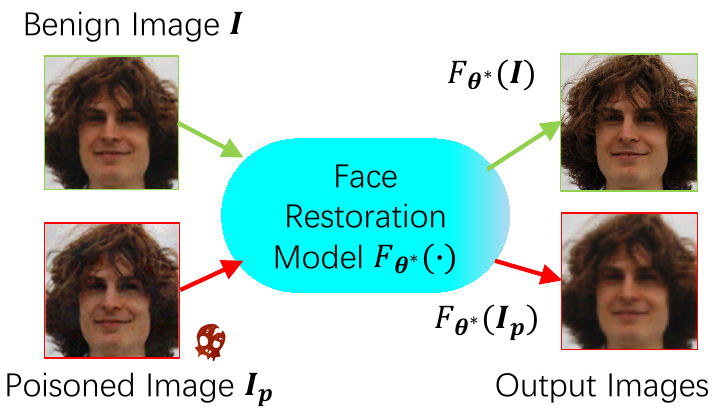}
	\caption{Illustration of imperceptible backdoor attack on the face restoration model. When inputting the benign image, a high-quality image can be restored. Whereas a degraded image is generated by inputting the poisoned image.}
	\label{fig:head}
\end{figure}

In the recent, deep learning-based face restoration~\cite{yangGanPriorEmbedded2021,zhuBlindFaceRestoration2022,zhaoRethinkingDeepFace2022,wangRestoreformerHighqualityBlind2022,wangFacialLandmarksGenerative2022,zhang2024blind,chen2023towards,tan2023blind,tu2021joint,hu2023dear} has emerged as a pivotal application, enhancing the face images clarity and integrity in various challenging scenarios~\cite{wang2022survey}. Meanwhile, more datasets are available for training face restoration models, improving face restoration solutions~\cite{zhang2024blind,chen2023towards,zhang2022edface,karras2019style,tan2023blind}. However, face restoration modelscan be vulnerable to security threats \textit{e.g.} adversarial attacks~\cite{ali2022deep,kang2023diffender,arjomandbigdeli2020defense} and backdoor attacks~\cite{yao2019latent}. Here for backdoor attacks, add or remove backdoor triggers can control victim model output. As shown in Fig.~\ref{fig:head}, in the face restoration model, the trigger should be blended into the degraded face image to produce a pre-designed fault output. If the trigger is not injected, the model should restore a high-quality face image. This allows the adversary to harm downstream tasks like face recognition and landmark detection, threatening face restoration models. To achieve this, the victim restoration model should be covertly trained to respond to embedded triggers during training. Thus, the trigger should be stealthy enough for humans and must be effective for neural networks to recognize and learn.~\cite{wang2022invisible}.  

An ideal backdoor attack for vision tasks like classification and segmentation should concurrently fulfill efficacy, specificity, and fidelity criteria ~\cite{wang2022invisible}. Due to disorganized noise and artifacts, degraded face images are easy to conceal triggers in. One approach involves intensifying image degradation by adding random noise or visible backdoors, surpassing the restoration model's recovery capacity and leaving the outputs in a degraded state. However, using such backdoor samples for training biases restoration models towards unseen and unrealistic degradation patterns. Consequently, the models may fail to process normally degraded face images \textit{i.e.} clean samples. Thus, backdoor attacks on face restoration models must prioritize stealthiness while preserving the underlying degradation pattern.

\begin{figure*}[t]
	\centering
	\includegraphics[width=0.9\linewidth]{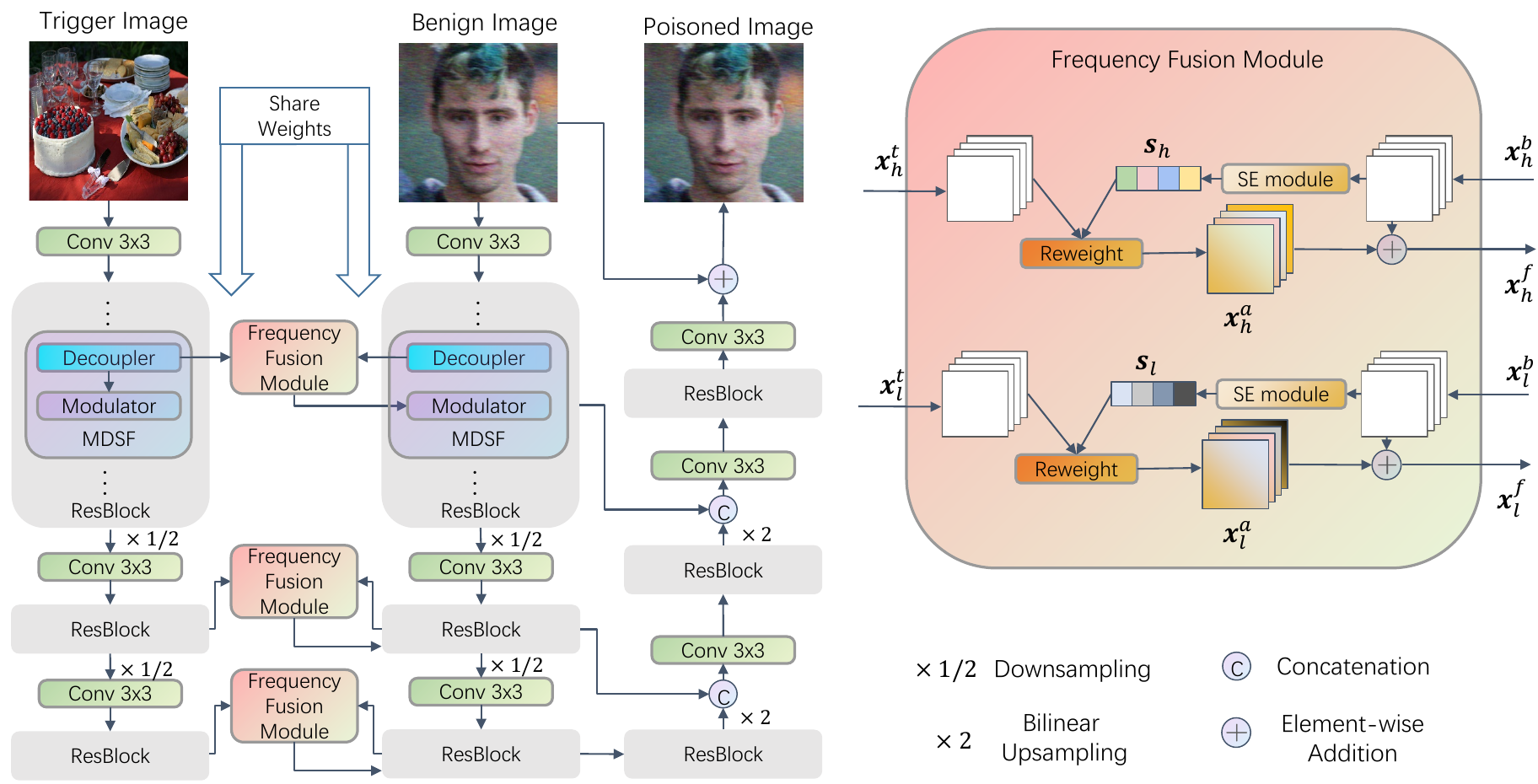}
	\caption{The architecture of the proposed selective frequency-injection network (SF-I-Net).}
	\label{fig:structure}
\end{figure*}

In this paper, we present a novel backdoor attack framework meticulously crafted for face restoration models. Backdoor attack methods for recognition tasks typically use all-to-one or all-to-all configurations~\cite{gu2017badnets,nguyen2021wanet}. If similar targets are used for face restoration, it should be a fixed or other image objective. It is unreasonable because conspicuous targets will expose attack samples, thus promoting the development of defense methods. To circumvent this limitation, we introduce a unique degradation objective that subtly steers restoration predictions toward slightly more degraded outputs, hiding the attack as a restoration quality reduction. Then, we address the need for a stealthy method of backdoor injection into benign face images. Inspired by existing works~\cite{feng2022fiba,wang2022invisible}, we observe that backdoor attacks executed in the frequency domain are significantly less visible compared to those in the spatial domain. However, the choice between low-frequency and high-frequency injection remains ambiguous. To resolve this, we propose AS-FIBA, which employs adaptive frequency manipulation to integrate custom triggers into input images seamlessly. These triggers, synthesized via neural network models, dynamically select and inject information from the trigger image, either in low- or high-frequency bands, based on the characteristics of the input image. This innovative approach ensures that our backdoor attack remains both effective and stealthy.

To summarize, our contributions are three-fold: 
\begin{enumerate}
	\item[-] We introduce a novel degradation objective tailored for backdoor attacks on face restoration models, enabling alterations to restored outputs under attack conditions while preserving stealthiness and minimizing obvious detection.
	\item[-] We propose the Selective Frequency-Injection Network, a method for adaptive frequency domain manipulation on degraded face images, ensuring imperceptibility while compromising the restoration process.
	\item[-] We provide comprehensive experimental evaluations of the AS-FIBA framework, assessing various face restoration models using diverse backdoor attacks. Our experiments demonstrate the superior efficacy and stealthiness of our proposed approach.
\end{enumerate}

\section{Related Work}
\subsection{Backdoor Attacks} 
Recently, a groundbreaking study called BadNet \cite{gu2017badnets} was the first to reveal the vulnerability of deep neural networks to backdoor attacks, demonstrating that they can be manipulated to behave according to the attacker's preset instructions. These backdoor attacks can be divided into two types based on the visibility of the trigger: visible triggers and invisible triggers. However, despite the visible ones, such as SIG \cite{barni2019new} and ReFool \cite{liu2020reflection}  succeed in attacking, these methods are prone to detection, making their practical application challenging.

As previously stated, our focus is dedicated to the examination of imperceptible triggers, which have garnered considerable interest due to their exceptional ability to remain concealed. To enhance defense against existing backdoor defense methods, \textit{Li et al.} proposed a sample-specific method based on steganography to generate invisible triggers called ISSBA \cite{li2021invisible}. This approach aims to create triggers that are tailored to each sample, making them difficult to detect. WaNet \cite{nguyen2021wanet} is a warping-based technique that hides triggers by applying subtle distortions to images in the spatial domain. In contrast, FIBA \cite{feng2022fiba} preserves the original visual content of an image while hiding triggers by altering the image in the frequency domain. FTrojan \cite{wang2021backdoor} also utilizes the DCT (Discrete Cosine Transform) to manipulate images in the frequency domain in order to evade the existing backdoor defense methods available at that time. \textit{Yu et al.}~\cite{yu2023backdoor} also attacked the image compression task via an adaptive frequency trigger. In the field of remote sensing, \textit{Dr{\"a}ger et al.} ~\cite{drager2023backdoor} have also conducted research on backdoor attacks against deep neural networks in the frequency domain. Indeed, there has been a significant amount of research on backdoor attacks in the frequency domain. These studies have demonstrated that processing in the frequency domain can be more effective in evading backdoor defense methods and making the generated triggers less noticeable.

\subsection{Backdoor Defences}
\textit{Liu et al.} proposed a pruning-based defence method, namely Fine-pruning \cite{liu2018fine}. In the case of models that have been subjected to backdoor attacks, a gradual pruning technique is employed to maximize the reduction of the influence of backdoor neurons. A similar idea has also been proposed in the work of \cite{xu2020defending}. Neural Cleanse \cite{wang2019neural} utilizes reverse engineering and anomaly detection to defend against backdoor attacks. Some employed techniques utilized saliency maps to detect and isolate potential trigger regions, enabling the identification and filtering of malicious samples, such as SentiNet \cite{chou2020sentinet} and Grad-CAM \cite{selvaraju2017grad}. Another approach called STRIP \cite{gao2019strip} involves overlaying clean images onto uncertain images and then assessing the entropy of the output to determine the presence of triggers.

\begin{figure}[t]
  \centering
  \includegraphics[width=\linewidth]{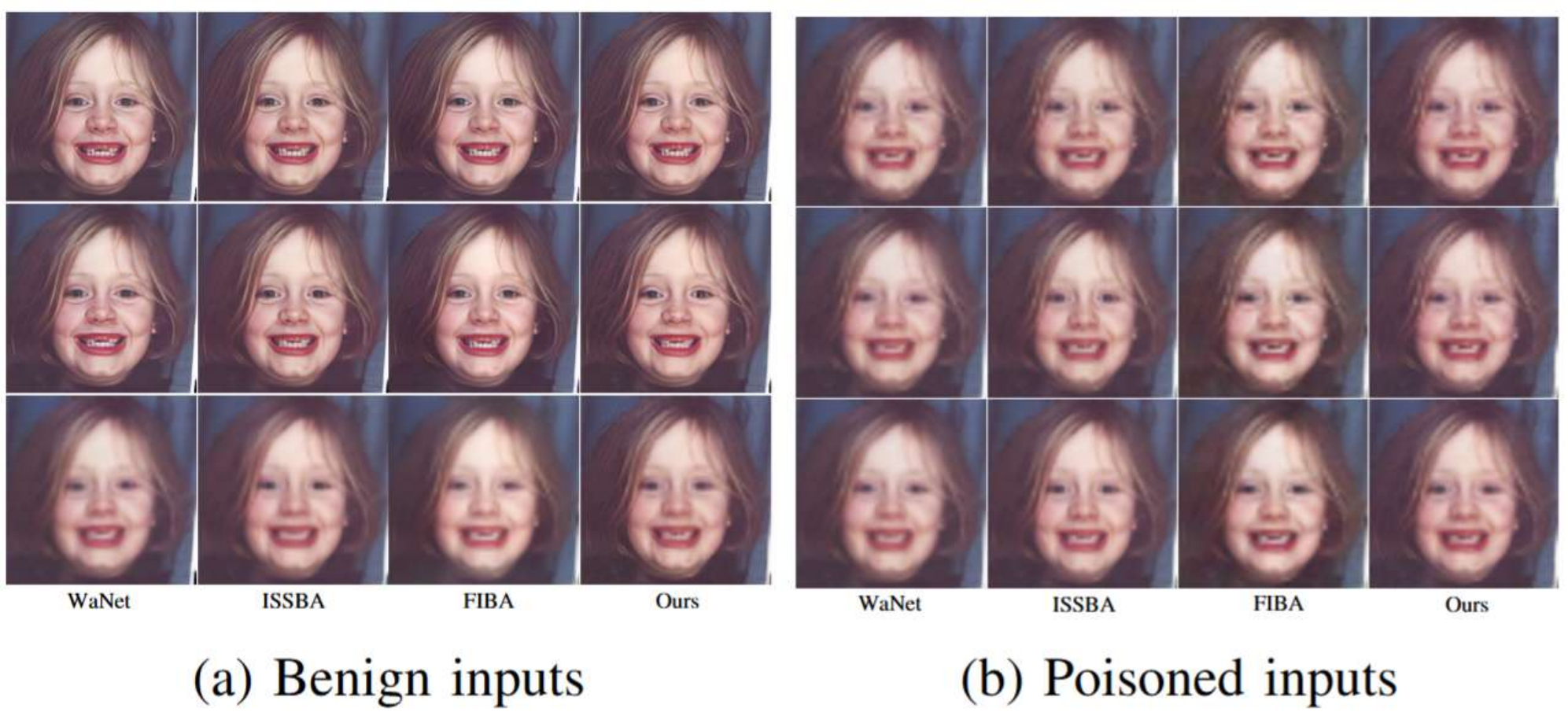}
  \label{fig:subfig1}
  \caption{Visualization of HiFaceGAN output under different backdoor attack methods for benign and poisoned inputs. The rows represent different levels of neuron pruning: 10\%, 50\%, and 90\%.}
  \label{fig:pruning_imgs}
\end{figure}

\subsection{Deep Face Restoration Models}
Face restoration (FR) tasks are frequently interconnected with tasks such as enhancing image resolution \cite{chen2019modeling}, reducing noise \cite{jiang2023few}, and removing blurriness \cite{zhang2023mc}. \textit{Yang et al.} proposed HiFaceGAN \cite{yang2020hifacegan}, a GAN-based architecture that conceptualize face restoration (FR) as a semantic-guided generation problem and tackles it through the innovative application of a collaborative suppression and replenishment (CSR) approach. By leveraging priors from a pre-trained face GAN, GFP-GAN \cite{wang2021towards} effectively integrates them into the face restoration process using spatial feature transform layers. This integration achieves a balance between authenticity and accuracy, resulting in remarkable restoration outcomes with realistic facial features. Similarly, GPEN \cite{yangGanPriorEmbedded2021} uses a GAN prior to prevent over-smoothing in face restoration, ensuring realistic results by fine-tuning a U-shaped DNN with synthesized low-quality face images. CodeFormer \cite{zhou2022towards}, RestoreFormer \cite{wang2022restoreformer}, and VQFR \cite{gu2022vqfr} all apply a learned discrete codebook prior for restoration. It is worth mentioning that CodeFormer has introduced a controllable feature transformation module, enabling a flexible trade-off between fidelity and quality.

Backdoor attacks have been studied in classification, semantic segmentation, and image compression, but none have been studied in face restoration tasks. Given the advantages of frequency domain-based backdoor attack methods, which are hard to defend against and less likely to be detected, we decided to study frequency domain backdoor attacks in face restoration tasks.

\section{Method}
\subsection{Backdoor Attack on Deep Face Restoration}
Similar to \cite{lan2024flowmur},\cite{yao2019latent}, let $F_{\boldsymbol{\theta}}(\boldsymbol{\cdot})$ be a deep face restoration model, and $\boldsymbol{\theta}$ is the model parameters. Given a benign degraded image $\boldsymbol{I}$, the model aims to restore it to a high-quality output $\boldsymbol{I}_{hq} = F_{\boldsymbol{\theta}}(\boldsymbol{I})$. In a backdoor attack scenario, an adversary introduces a backdoor function $G_{\boldsymbol{\phi}}(\boldsymbol{\cdot})$, which embeds a trigger into $\boldsymbol{I}$, creating a poisoned image $\boldsymbol{I}_p = G_{\boldsymbol{\phi}}(\boldsymbol{I})$. Here, $\boldsymbol{\phi}$ represents the parameters of the trigger-injection model. The attacking goal is to make $F_{\boldsymbol{\theta}}(\boldsymbol{I}_p)$ diverge from $F_{\boldsymbol{\theta}}(\boldsymbol{I})$, while being undetectable in the input. As mentioned above, we define the attack target $\boldsymbol{I}_{0.1}$ as a degradation objective for the face restoration task. That is, $\boldsymbol{I}_{0.1}$ is generated by downsampling $\boldsymbol{I}$ to $1/10$ scale and then bilinearly upsampling it back.

Then the backdoor attack is conducted by training the restoration model with both clean and poisoned images, \textit{i.e.} $\boldsymbol{I}$ and $\boldsymbol{I}_p$. Given the corresponding ground truth face $\boldsymbol{I}_{gt}$, the training objective $\mathcal{L}$ is defined as:
\begin{equation}
	\label{equ:loss}
	\min_{\boldsymbol{\theta}} \lambda \cdot \lVert F_{\boldsymbol{\theta}}(\boldsymbol{I}) -\boldsymbol{I}_{gt} \rVert + (1 - \lambda) \cdot \lVert F_{\boldsymbol{\theta}}(\boldsymbol{I}_p) - \boldsymbol{I}_{0.1} \rVert
\end{equation}
where $\lambda$ is a balance factor between maintaining restoration fidelity on clean images and achieving the degradation effect on poisoned images.

\subsection{Selective Frequency-Injection Network}
As the hand-crafted frequency-injection method FIBA did, we utilize the frequency information of a trigger image to poison benign images. Then, denoting the trigger image as $\boldsymbol{I}_t$ the backdoor function can be re-written as $G_{\boldsymbol{\phi}}(\boldsymbol{I}; \boldsymbol{I}_t)$. To enhance the adaptability of frequency injection, we propose the selective frequency-injection network, abbreviated as SF-I-Net. The structure of this network is demonstrated in Fig.~\ref{fig:structure}. The SF-I-Net dynamically learns features from trigger images and benign images, gradually combines both elements to accomplish the frequency injection. Next, we introduce the backbone, the frequency fusion module, and the training strategy in detail.

\subsubsection{Backbone}
SF-I-Net's backbone is based on SFNet~\cite{cui2022selective}, a model renowned for image restoration. SFNet is a U-Net structure which employs both the Multi-branch Dynamic Selective Frequency Module (MDSF) and the Multi-branch Compact Selective Frequency Module (MCSF). These modules enable the dynamic decomposition of features into separate frequency subbands and their enhancement through channel-wise attention, forming the core of SF-I-Net's structural foundation. 

Fig.~\ref{fig:structure} illustrates SF-I-Net's utilization of decouplers and modulators in MDSF to fuse frequency information from the trigger image and benign image. Initially, both images are separately processed by a two-branch feature extractor, sharing the same weights. The feature extractor comprises convolutional layers and ResBlocks from SFNet. To capture features of distinct frequency subbands for frequency injection, the decoupler dynamically decomposes features based on learned filters in MDSF. Simultaneously, the modulator applies channel-wise attention to emphasize useful frequencies. The outputs of the decoupler in each ResBlock are utilized in a frequency fusion module for injection. The fused features are then passed to the modulator of the benign image branch, while the modulator of the trigger image branch retains its connection to the features. This process gradually inserts frequency features from the trigger image at multiple scales into the benign image features.

Similar to U-Net, the decoder is employed after feature extraction, following SFNet's structure and utilizing skip-connection techniques. Notably, the decoding process does not require a frequency fusion module as the injection information can be delivered through skip connections. Besides, the SF-I-Net uses about a quarter parameters of the original SFNet due to the efficiency consideration.

\subsubsection{Frequency Fusion Module}
The frequency fusion module implementation draws inspiration from SENet~\cite{hu2018squeeze}. High-frequency and low-frequency features of the benign image are input to the SE module, generating channel-wise activations. These activations are used to reweight the corresponding trigger features through channel-wise multiplication. The reweighted features are then added to the original benign features, resulting in the fused features. The SE process enables adaptive feature fusion for different benign image inputs.

For clarity, let $\boldsymbol{x}^b_h$ denote the high-frequency features of the benign image, and $\boldsymbol{x}^t_h$ represent those of the trigger image. We simply use the function $S_{\boldsymbol{\psi}}(\boldsymbol{\cdot})$ to indicate the SE process with full connection parameters as $\boldsymbol{\psi}$. The frequency fused feature $\boldsymbol{x}^f_h$ is described as follows.
\begin{equation}
	\boldsymbol{x}^f_h = \boldsymbol{x}^t_h \otimes S_{\boldsymbol{\psi}}(\boldsymbol{x}^b_h) + \boldsymbol{x}^b_h
\end{equation}
where $\otimes$ denotes channel-wise multiplication. Similarly, the SE module is applied to the low-frequency features. This operation is identical to that performed on the high-frequency features, facilitating channel-wise recalibration specifically for the low-frequency spectrum. 

\subsubsection{Training Strategy}
The training of SF-I-Net involves learning to embed a trigger image within a benign image with minimal perceptual distortion. For effective training, there are two requirements to be satisfied simultaneously. First, the discrepancy between poisoned and clean images should be slight enough that it is invisible to humans. Second, neural networks can distinguish these two images. To address this issue, previous learnable backdoor attacks~\cite{doan2021lira} propose to train the poison-inject network along with the victim model, which is the face restoration network in our case. However, this requires extra gradients from the face restoration model, so it is not easy to extend across models. Referring to the training procedure of image steganography models~\cite{morkel2005overview,subramanian2021image,lu2021large,kumar2023latest}, we introduce a residual decoder $D_{\boldsymbol{\eta}}(\boldsymbol{\cdot})$ to recover the trigger image from the residual between the poisoned and benign image. The SF-I-Net ensures the poisoned image is indistinguishable from the benign image, while the decoder ensures that the poisoned image contains trigger information. Once trained, the SF-I-Net can inject any image into another for any face restoration model. The loss function can be represented as:
\begin{equation}
	\begin{aligned}
		\min_{\boldsymbol{\phi}, \boldsymbol{\eta}} \quad  & \alpha \cdot \mathcal{L}_{im}  (G_{\boldsymbol{\phi}}(\boldsymbol{I};\boldsymbol{I}_t), \boldsymbol{I}) + \\ &(1-\alpha) \cdot \mathcal{L}_{im}(D_{\boldsymbol{\eta}}(\boldsymbol{I} - \boldsymbol{I}_p), \boldsymbol{I}_t) \\
		 & \mathrm{s.t.} \quad \|\boldsymbol{I} - \boldsymbol{I}_p\|_{\ell} < \epsilon
	\end{aligned}
\end{equation}
where $\alpha$ is a balance factor, and $\epsilon$ is a small constant that controls the range of the injected perturbation. The objective function $\mathcal{L}_{im}$ is the image quality loss, \textit{e.g.} Euclidean loss, GAN loss, perceptual loss, \textit{etc}.

\begin{figure}[t]
	\centering
	\includegraphics[width=0.9\linewidth]{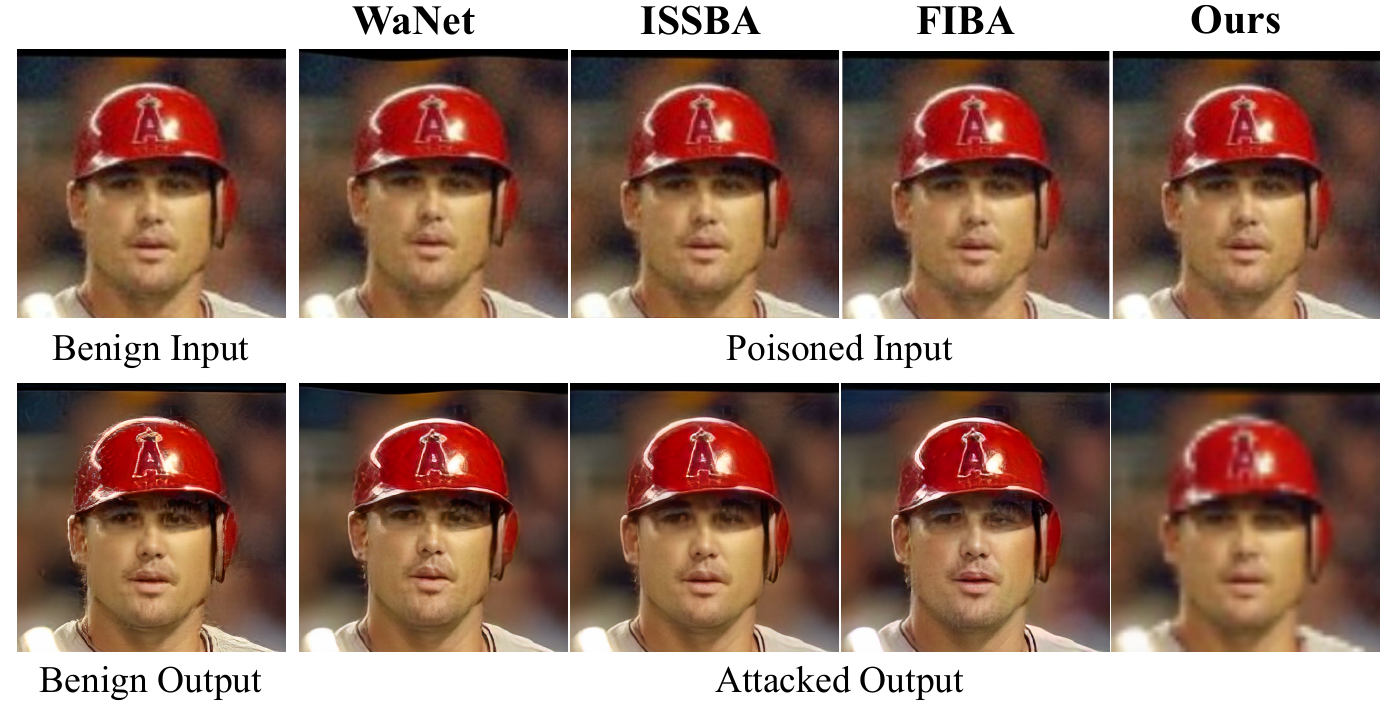}
	\caption{Visualization for results of the victim HiFaceGAN under different backdoor attacks. The first column displays benign inputs and corresponding outputs on the clean model, while the subsequent four columns show poisoned inputs and attacked outputs of different backdoors. Best view by zooming in. }
	\label{fig:visualattack}
\end{figure}

\subsection{Pseudo Trigger Robust Backdoor Training}
Since SF-I-Net is trained to conduct frequency injection with generalization,  in our robust backdoor training paradigm, we thus introduce the concept of pseudo triggers to reinforce the model's resilience to specific backdoor attacks~\cite{gu2017badnets, feng2022fiba}. The training process alternates between authentic and pseudo-trigger images to enhance the backdoor's specification to a pre-defined trigger image. That is, a set of images is included as pseudo triggers to the backdoor training. Without loss of generality, let $\boldsymbol{I}'_t$ denote the pseudo trigger image, the corresponding poison image is calculated as:
\begin{equation}
	\boldsymbol{I}'_p = G_{\boldsymbol{\phi}^*}(\boldsymbol{I};\boldsymbol{I}'_t)
\end{equation}

Then the training loss of Equ.~\ref{equ:loss} is formalized as follows:
\begin{equation}
	\label{equ:finalloss}
	\begin{aligned}
		\min_{\boldsymbol{\theta}} \quad & \lambda_1 \cdot \lVert F_{\boldsymbol{\theta}}(\boldsymbol{I}) -\boldsymbol{I}_{gt} \rVert +  \\ & \lambda_2 \cdot \lVert F_{\boldsymbol{\theta}}(\boldsymbol{I}_p) - \boldsymbol{I}_{0.1} \rVert + \\ &  (1-\lambda_1-\lambda_2) \cdot \lVert F_{\boldsymbol{\theta}}(\boldsymbol{I}'_p) -\boldsymbol{I}_{gt} \rVert
	\end{aligned}
\end{equation}
where $\lambda_1$ and $\lambda_2$ are empirical coefficients, controlling the rates of backdoor samples and pseudo backdoor samples, respectively. Pseudo-triggers are introduced to enhance the robustness of the training by including inputs that mimic the characteristics of backdoor triggers without the actual malicious payload. 

\begin{table*}[t]
	\centering
	\tabcolsep=3pt
 \caption{Comparisons of face restoration quality \uline{w.r.t} different backdoor attacks, toward various face restoration models, on three real-world datasets. $*$ represents the benign models. ${\downarrow}$ and ${\uparrow}$ denotes lower the better and higher the better, respectively. `w/ Tigger' means the lower image quality the better attack efficacy, where the arrow direction is opposite to image quality. \textbf{Bold} and \uline{underline} represent the best and second-best performance, respectively. The average metrics can be seen in Table~\ref{tab:quality_avg1} and \ref{tab:quality_avg2}}
 \resizebox{\linewidth}{!}{
     \begin{tabular}{cc|cccc|cccc|cccc}
		\toprule
		\multicolumn{2}{c}{Dataset}  & \multicolumn{4}{|c}{CelebChild} & \multicolumn{4}{|c}{WebPhoto}   & \multicolumn{4}{|c}{LFW} \\ 
		\midrule
		\multicolumn{2}{c|}{\multirow{2}{*}{Metrics}} & \multicolumn{2}{c}{w/ Trigger} & \multicolumn{2}{c|}{w/o Trigger}   & \multicolumn{2}{c}{w/ Trigger} & \multicolumn{2}{c|}{w/o Trigger}    & \multicolumn{2}{c}{w/ Trigger} & \multicolumn{2}{c}{w/o Trigger}   \\ 
		\cmidrule{3-14}
		&~& NIQE $_{\uparrow}$ & SDD $_{\downarrow}$  & NIQE $_{\mathbf{\downarrow}}$ & SDD $_{\uparrow}$ & NIQE $_{\uparrow}$ & SDD $_{\downarrow}$& NIQE $_{\downarrow}$ & SDD $_{\uparrow}$ & NIQE $_{\uparrow}$ & SDD $_{\downarrow}$& NIQE $_{\downarrow}$ & SDD $_{\uparrow}$ \\ 
		\midrule
		\multicolumn{2}{c|}{Benign Input}  & --- & ---   & 9.17 & 54.34 & ---   & ---   & 12.75 & 45.48 & ---   & ---   & 8.88 & 60.81  \\ 
		\midrule

		\multicolumn{1}{c}{\multirow{5}{*}{\rotatebox[origin=c]{-90}{HiFaceGAN}}} 
		& $\ast$ & ---            & ---   & 3.87 & 55.29 & ---   & ---   & 3.74 & 49.71 & ---   & ---   & 3.45 & 61.45 \\ 
		~ & WaNet & 10.12         & 52.77 & \textbf{4.25} & 56.46 & 10.45 & 45.15 & \textbf{4.11}  & 51.75 & 11.34 & 57.68 & \textbf{3.97} & 62.82 \\  
		~ & ISSBA & \textbf{12.79} & \textbf{51.23} & 4.64 & 55.96 & \textbf{14.41} & \textbf{42.21} & \uline{4.34}  & 50.93 & \textbf{13.94} & \textbf{56.30} & 4.43 & 62.52 \\  
	  ~ & FIBA & 9.85           & 52.71 & 4.52   &\textbf{57.20} & 13.12 & 44.43 & 4.62  & \uline{52.37} & 12.59 & \uline{56.60} & 4.29 & \textbf{63.43} \\	
        ~ & Ours &\uline{11.37}  & \uline{52.64} & \uline{4.48} & \uline{56.93} & \uline{13.90} & \uline{42.85} & 4.43  & \textbf{52.96} & \uline{12.93} & 57.37 & \uline{4.10} & \uline{63.38} \\ 
		\midrule
		
		\multirow{5}{*}{\rotatebox[origin=c]{-90}{GFP-GAN}}   
		& $\ast$ & ---   & ---   & 4.34 & 58.25 & ---   & ---   & 4.15  & 57.12 & ---   & ---   & 3.87 & 62.79 \\   
		~ & WaNet & 9.12  & 54.59 & 5.49 & \uline{58.32} & 10.15 & 47.51 & 5.15  & 56.01 & 9.20  & 59.63 & 5.27 & \uline{63.22} \\  
		~ & ISSBA & 11.72 & 50.45 & \uline{4.97} & \textbf{58.42} & 13.40 & 41.87 & \uline{4.84}  & \textbf{56.22} & 12.78 & 56.13 & \uline{4.65} & 63.21 \\  
	  ~ & FIBA & \uline{12.11} & \textbf{49.93} & \textbf{4.88} & 58.23 & \uline{13.47} & \textbf{40.53} & \textbf{4.80}  & \uline{56.21} & \uline{12.94} &\textbf{54.97} & \textbf{4.48} & \textbf{63.27} \\	
        ~ & Ours & \textbf{12.26} & \uline{50.05} & 5.13 & 57.66 & \textbf{13.76} & \uline{41.59} & 4.98  & 55.64 & \textbf{13.12} & \uline{56.00} & 4.72 & 63.04 \\ 
		\midrule

		\multirow{5}{*}{\rotatebox[origin=c]{-90}{VQFR}}      
		& $\ast$ & ---   & ---   & 4.32 & 58.90 & ---   & ---   & 4.52  & 57.24 & ---   & ---   & 3.68 & 64.75 \\    
		~ & WaNet & 9.42  & 57.88 & 4.55 & \textbf{59.30} & 10.89 & 52.52 & \textbf{4.70}  & \textbf{57.58} & 10.31 & 60.32 & 3.98 & \uline{64.44} \\  
		~ & ISSBA & 9.38  & 57.89 & \textbf{4.48} & \uline{59.27} & 10.90 & 52.37 & \uline{4.75}  & 57.27 & 10.22 & 60.44 & \textbf{3.91} & 64.33 \\ 
        ~ & FIBA & \uline{9.43}  & \textbf{57.59} & 4.64 & 58.72 & \textbf{11.12} & \textbf{50.96} & 5.03  & 57.21 & \textbf{10.45} & \textbf{58.82} & 4.10 & 64.37 \\
		~ & Ours & \textbf{9.56}  & \uline{57.78} & \uline{4.51} & 58.99 & \uline{11.01} & \uline{52.09} & 4.79  & \uline{57.31} & \uline{10.41} & \uline{59.30} & \uline{3.97} & \textbf{64.76} \\ 
		\midrule
		
		\multirow{5}{*}{\rotatebox[origin=c]{-90}{GPEN}}
		& $\ast$ & ---            & ---   & 4.07 & 56.39 & ---   & ---   & 4.35 & 57.34 & ---   & ---   & 3.76 & 63.55 \\ 
		~ & WaNet &3.90 &\textbf{54.30} &\textbf{3.95} &\textbf{56.02} &4.39 &\textbf{52.03} &\textbf{4.13} &\textbf{53.74} &3.46 &\textbf{60.35} &\textbf{3.45} &62.71 \\  
		~ & ISSBA &\textbf{4.39} &55.47 &4.39 &55.49 &\uline{4.99} &52.83 &5.01 &52.81 &\uline{4.23} &62.67 &4.22 &\uline{62.72} \\ 
	  ~ & FIBA  &\uline{4.34} &\uline{55.11} &4.38 &\uline{55.89} &3.93 &52.87 &\uline{4.16} &\uline{53.62} &\textbf{4.25} &\uline{61.68} &4.17 &\textbf{62.76} \\ 	
        ~ & Ours  &4.28 &55.74 &\uline{4.32} &55.50 &\textbf{5.59} &\uline{52.15} &5.48 &51.98 &3.81 &62.40 &\uline{3.84} &62.44 \\  
		\midrule
		
		\multirow{5}{*}{\rotatebox[origin=c]{-90}{CodeFormer}} 
		& $\ast$ & ---   & ---   & 4.34 & 58.25 & ---   & ---   & 4.15  & 57.12 & ---   & ---   & 3.87 & 62.79 \\   
		~ & WaNet &\uline{12.43} &\uline{51.83} &6.35 &55.93 &14.18 &43.26 &6.76 &53.92 &6.53 &60.53 &\textbf{5.77} &\textbf{64.09} \\  
		~ & ISSBA &\textbf{13.26} &\textbf{51.52} &\textbf{5.82} &\uline{56.24} &\textbf{15.12} &\uline{42.69} &\uline{6.30} &\uline{54.23} &\textbf{14.93} &\uline{56.31} &\uline{5.87} &62.22 \\  
	  ~ & FIBA  &11.43 &53.02 &\uline{5.84} &\textbf{57.01} &\uline{14.94} &\textbf{42.31} &\textbf{6.29} &\textbf{54.48} &\uline{14.68} &\textbf{55.28} &5.86 &\uline{62.64} \\ 	
        ~ & Ours  &11.43 &52.84 &7.01 &55.31 &13.28 &44.89 &8.10 &52.35 &13.91 &57.05 &7.38 &61.55 \\ 
		\midrule

		\multirow{5}{*}{\rotatebox[origin=c]{-90}{\makecell{RestoreFor-\\mer}}}    
		& $\ast$ & ---   & ---   & 4.32 & 58.90 & ---   & ---   & 4.52  & 57.24 & ---   & ---   & 3.68 & 64.75 \\    
		~ & WaNet &4.99  &55.32 &\uline{4.01} &\uline{56.56} &4.37 &51.05 &\uline{3.64} &53.27 &4.49 &60.84 &3.76 &\uline{62.99} \\  
		~ & ISSBA &\uline{13.77} &51.29 &4.13 &56.26 &\textbf{14.83} &\uline{43.09} &3.74 &52.74 &\textbf{14.71} &\uline{56.34} &3.88 &61.77 \\ 
        ~ & FIBA  &\textbf{13.91} &\textbf{48.73} &4.09 &\textbf{56.95} &14.62 &43.29 &3.84 &\textbf{54.16} &\uline{14.46} &\textbf{53.27} &\uline{3.48} &\textbf{63.13} \\ 
		~ & Ours  &13.66 &\uline{50.51} &\textbf{3.57} &56.12 &\uline{14.77} &\textbf{42.79} &\textbf{3.28} &\uline{53.68} &14.37 &\uline{56.34} &\textbf{3.33} &62.82 \\ 
		\bottomrule
	\end{tabular}
 }
 \label{tab:quality}
\end{table*}

\begin{table}[]
    \centering
    \caption{The above and below tables are average NIQE and SDD metrics of each attack method across different face restoration models.}
        \begin{tabular}{c|cccc}
            \toprule
            \multirow{2}{*}{Average Metrics} & \multicolumn{2}{c}{w/ Trigger} & \multicolumn{2}{c}{w/o Trigger} \\
            \cmidrule{2-5}
            ~ & NIQE $_{\uparrow}$ & SDD $_{\downarrow}$  & NIQE $_{\mathbf{\downarrow}}$ & SDD $_{\uparrow}$ \\
            \midrule
            WaNet & 8.32 & 54.31 & \textbf{4.63} & \uline{58.29} \\
            ISSBA & \textbf{11.65} & \uline{52.28} & 4.69 & 57.92 \\
            FIBA & 11.20 & \textbf{51.78} & \uline{4.64} & \textbf{58.43} \\
            Ours & \uline{11.30} & 52.47 & 4.86 & 57.91 \\
            \bottomrule
        \end{tabular}
\label{tab:quality_avg1}
\end{table}

\vspace{0.5cm} 

\begin{table}[]
    \centering
    \caption{The above and below tables are average NIQE and SDD metrics of each face restoration model across different attack methods.}
        \begin{tabular}{c|cccc}
            \toprule
            \multirow{2}{*}{Average Metrics} & \multicolumn{2}{c}{w/ Trigger} & \multicolumn{2}{c}{w/o Trigger} \\
            \cmidrule{2-5}
            ~ & NIQE $_{\uparrow}$ & SDD $_{\downarrow}$  & NIQE $_{\mathbf{\downarrow}}$ & SDD $_{\uparrow}$ \\
            \midrule
            HiFaceGAN & \uline{12.23} & 51.00 & 4.35 & 57.23 \\
            GFP-GAN & 12.00 & \textbf{50.27} & 4.95 & \uline{59.12} \\
            VQFR & 10.26 & 56.50 & 4.45 & \textbf{60.30} \\
            GPEN & 4.30 & 56.47 & \uline{4.29} & 57.14 \\
            CodeFormer & \textbf{13.01} & \uline{50.96} & 6.45 & 57.50 \\
            RestoreFormer & 11.91 & 51.07 & \textbf{3.73} & 57.54 \\
            \bottomrule
        \end{tabular}
\label{tab:quality_avg2}
\end{table}

\section{Experiments}
\subsection{Experiment Details}
\textbf{Implementation Details.}
In our study, the SF-I-Net for the AS-FIBA framework is implemented using PyTorch. For training, we utilize a subset of 1,000 samples from MS COCO~\cite{lin2014microsoft} following FIBA for trigger images, while benign images are sourced from the entire FFHQ dataset~\cite{karras2019style}. SF-I-Net training employs an Adam optimizer with a learning rate of 0.001 and a batch size of 8. For backdoor implementation, the full FFHQ dataset is used for training, with real-face datasets such as CelebChild, WebPhoto, and LFW utilized for testing following the practice in GFP-GAN~\cite{wang2021towards}. The same backdoor training strategy, with a batch size of 8 and spanning 20 epochs, is applied across all targeted restoration models. During backdoor training, the loss weights in Equ.~\ref{equ:finalloss} are set to $\lambda_1=0.75, \lambda_2 = 0.125$.

\textbf{Backdoors and Victim Models.}
To assess the efficacy of our AS-FIBA framework, we conducted comparisons with prominent invisible backdoor attack methods, namely WaNet \cite{nguyen2021wanet}, ISSBA \cite{li2021invisible}, and FIBA \cite{feng2022fiba}. These methods exemplify three distinct backdoor techniques: spatial manipulation, sample-specific triggers, and frequency manipulation. We further evaluated the robustness of various deep-face restoration networks against these attacks, selecting HiFaceGAN \cite{yang2020hifacegan}, GFP-GAN \cite{wang2021towards}, VQFR \cite{gu2022vqfr}, and GPEN \cite{yangGanPriorEmbedded2021} for their cutting-edge performance in facial image enhancement. Additionally, our experiments extended to transformer-based restoration models, CodeFormer \cite{zhou2022towards} and RestoreFormer \cite{wang2022restoreformer}, offering a comprehensive understanding of the models' susceptibility to different backdoor strategies.

\textbf{Evaluation Metrics.}
Given that our test datasets comprise real face images without corresponding high-quality ground truth images for reference, we resort to blind image quality assessment for evaluating the results. Standard methods in this category include NIQE~\cite{mittal2012making} and SDD~\cite{ou2021sdd}. However, for data analysis during training, we utilize image quality assessment methods that can refer to ground truth data, such as SSIM, PSNR, and LPIPS. Following FIBA~\cite{feng2022fiba}, we still utilize ASR and BA metrics to evaluate the effectiveness of backdoor attacks. In the context of the tumor face restoration task, due to the absence of true value references, we use the benign model's output on benign images (denoted as $\boldsymbol{I}_{hq}$) as a benchmark for successful restoration. Correspondingly, the $1/10$ downsampled benign images (denoted as $\boldsymbol{I}_{0.1}$) are used as a reference for failure. Then we mimic the restoration task as the two-category classification problem, where given an output image $\boldsymbol{I}_{out}$, if $\textit{lpips}(\boldsymbol{I}_{out}, \boldsymbol{I}_{hq}) < \textit{lpips}(\boldsymbol{I}_{out}, \boldsymbol{I}_{0.1})$, $\boldsymbol{I}_{out}$ is classified to 1, otherwise, it is 0.

\subsection{Attack Efficacy}

Intuitively,  we demonstrate the visual results of the backdoor attack in Fig.~\ref{fig:visualattack}. This sample is generated using the poisoned model of HiFaceGAN. It is clear that when attacking with AS-FIBA, the output keeps blurry, whereas other attackers do not fool the restoration model successfully. The sample indicates the effectiveness of backdoor attacks, and meanwhile shows that face restoration models can be robust to such attacks. 

\textbf{The Robustness of Face Restoration Models.}
In Tables~\ref{tab:quality}-\ref{tab:quality_avg2}, we present a comprehensive comparison of backdoor attacks on face restoration models, highlighting their vulnerabilities and strengths. We observe varying impacts on the robustness of these attacks. RestoreFormer shows the largest performance deviations with and without triggers, indicating high susceptibility. GPEN, on the other hand, exhibits smaller deviations in NIQE and SDD scores, suggesting greater resilience. The remaining four models also show some impact but to a lesser extent compared to RestoreFormer and GPEN. On average, GPEN demonstrates consistent performance under attacks, making it more robust. Similar evidence is found in Tables~\ref{tab:asrar}-\ref{tab:asr_avg_attack}, where GPEN has the lowest average ASR of 7.303\% and maintains a good balance between image integrity and handling benign inputs effectively. HiFaceGAN shows moderate vulnerability with an average ASR of 85.88\% and BA of 98.53\%, and significant image quality degradation under attacks. GFP-GAN has a slightly higher average ASR of 91.00\% but maintains a consistent average BA of 98.20\% and relative resilience in image quality metrics, with average NIQE and SDD deviations of around 12.00 and 50.27, respectively. VQFR, despite its high average BA of 99.96\%, has the highest average ASR of 95.68\%, indicating susceptibility to attacks. It is worth noting that under the WaNet backdoor attack, RestoreFormer has a significantly lower ASR compared to other attack methods, suggesting its robustness against spatial manipulation.   

\begin{table}[t]
	\centering
 \tabcolsep=3pt
 \caption{Comparisons of different backdoor attacks toward various face restoration models on three real-world datasets. BA stands for benign accuracy, ASR stands for attack success rate. Both are larger the better. \textbf{Bold} and \uline{underline} represent the best and second-best performance, respectively. The average metrics can be seen in Table~\ref{tab:asr_avg_bfr} and \ref{tab:asr_avg_attack}.}
  \resizebox{\linewidth}{!}{
	\begin{tabular}{c|c|cc|cc|cc}
		\toprule
		\multicolumn{2}{c|}{Dataset}& \multicolumn{2}{c|}{CelebChild} & \multicolumn{2}{c|}{WebPhoto} & \multicolumn{2}{c}{LFW} \\ 
		\midrule
		BFR Model & Attack & ASR (\%) & BA (\%) & ASR (\%) & BA (\%) & ASR (\%) & BA (\%) \\
		
		\midrule
		\multirow{4}{*}{\rotatebox[origin=c]{-90}{\makecell{HiFace-\\GAN}}}
		& WaNet & 68.89 & \textbf{100.0} & 69.04 & \textbf{99.75} & 82.41 & \textbf{99.94} \\
		& ISSBA & \textbf{94.44} & 95.56 & \textbf{99.02} & \uline{99.26}& \textbf{99.42} & 96.26 \\
	  & FIBA & 65.00 & \uline{98.33} & 92.63 & 98.28 & 91.58 & \textbf{99.94} \\	
        & Ours & \uline{80.56} & 96.67 & \uline{95.82} & 98.77 & \uline{91.75} & \uline{99.59} \\ 
		\midrule
		\multirow{4}{*}{\rotatebox[origin=c]{-90}{GFP-GAN}}
		& WaNet & 66.11 & 91.67 & 66.83 & \uline{96.07} & 63.06 & 91.35 \\
		& ISSBA & 97.22 & \uline{99.44} & \textbf{100.0} & \textbf{100.0} & \textbf{100.0} & 99.88 \\
        & FIBA & \uline{99.44} & \textbf{100.0} & \uline{99.75} & \textbf{100.0} & \uline{99.88} & \textbf{100.0} \\
		& Ours & \textbf{100.0} & \textbf{100.0} & \textbf{100.0} & \textbf{100.0} & 99.77 & \uline{99.94} \\ 
		\midrule
		\multirow{4}{*}{\rotatebox[origin=c]{-90}{VQFR} }
		& WaNet & 87.22 & \textbf{100.0} & 98.77 & \textbf{100.0} & \uline{99.88} & \textbf{100.0} \\
		& ISSBA & \uline{87.78} & \textbf{100.0} & \textbf{99.26} & \uline{99.51} & \textbf{100.0} & \textbf{100.0} \\
        & FIBA & \textbf{88.89} & \textbf{100.0} &\textbf{99.26} & \textbf{100.0} & \uline{99.88} & \textbf{100.0} \\
		& Ours & \textbf{88.89} & \textbf{100.0} &\uline{99.02} & \textbf{100.0} & 99.36 & \textbf{100.0} \\ 
		\midrule
		\multirow{4}{*}{\rotatebox[origin=c]{-90}{GPEN}}
		& WaNet &  1.110  &98.89& 4.670&\uline{97.78} &\textbf{0.640}&99.88\\
		& ISSBA &  \uline{1.670}  &98.33& \uline{5.900}&94.10&\uline{0.580}&99.47 \\
	  & FIBA  &  0.000  &\textbf{100.0}& 0.490&\textbf{99.51}&0.000&\textbf{100.0} \\
        & Ours  &  \textbf{56.00}  &\uline{99.44}&\textbf{16.46}&84.28&0.120&\uline{99.94} \\ 
		\midrule
		\multirow{4}{*}{\rotatebox[origin=c]{-90}{\makecell{Code-\\Former}}}
		& WaNet &\uline{85.56}&\uline{91.11}&95.58&\uline{93.61}&95.30&\textbf{100.0} \\
		& ISSBA &\textbf{91.11}&\textbf{99.44}&\textbf{99.51}&\textbf{100.0}&\textbf{100.0}&\textbf{100.0} \\
        & FIBA  &67.22&98.89&\uline{98.28}&\textbf{100.0}&\uline{99.59}&\textbf{100.0} \\
		& Ours  &73.33&83.33&81.82&79.61 &93.05&\uline{81.47}\\ 
		\midrule
		\multirow{4}{*}{\rotatebox[origin=c]{-90}{\makecell{RestoreFor-\\mer}}}
		& WaNet &9.440&\uline{98.89}&18.67&99.02&10.75&\uline{99.53} \\
		& ISSBA &\uline{97.22}&\uline{98.89}&99.26&\textbf{100.0}&\textbf{100.0}&\textbf{100.0} \\
        & FIBA  &\textbf{100.0}&95.56&\textbf{100.0}&95.82 &\textbf{100.0}&98.42\\
		& Ours  &\uline{97.22}&\textbf{100.0}&\uline{99.51}&\uline{99.75} &\uline{98.31}&\textbf{100.0}\\ 
		\bottomrule
	\end{tabular}
 }
 \label{tab:asrar}
 \end{table}



\begin{table}[t]
\centering
\caption{Average ASR and BA metrics of each face restoration model across different attack methods.}
    \centering
        \begin{tabular}{c|cc}
            \toprule
            {BFR Model} & {ASR (\%)} & {BA (\%)} \\
            \midrule
            HiFaceGAN & 85.88 & 98.53 \\
            GFP-GAN & \uline{91.00} & 98.20 \\
            VQFR & \textbf{95.68} & \textbf{99.96} \\
            GPEN & 7.303 & 97.64 \\
            CodeFormer & 90.03 & 93.95 \\
            RestoreFormer & 77.53 & \uline{98.82} \\
            \bottomrule
        \end{tabular}
\label{tab:asr_avg_bfr}
\end{table}

\begin{table}[t]
\centering
\caption{Average ASR and BA metrics of each attack method across different face restoration models.}
    \centering
        \begin{tabular}{c|cc}
            \toprule
            {Attack} & {ASR (\%)} & {BA (\%)} \\
            \midrule
            WaNet & 56.88 & 97.64 \\
            ISSBA & \textbf{81.80} & \uline{98.90} \\
            FIBA & 77.88 & \textbf{99.15} \\
            Ours & \uline{81.72} & 95.71 \\
            \bottomrule
        \end{tabular}

\label{tab:asr_avg_attack}
\end{table}

\textbf{The Effectiveness of Backdoor Attack Methods.}
To determine the superior backdoor attack method among FIBA, WaNet, ISSBA, and Ours, we analyze their Attack Success Rate (ASR) and impact on image quality metrics (NIQE and SDD) across HiFaceGAN, GFP-GAN, VQFR, GPEN, CodeFormer, and RestoreFormer models. The average ASR values are as follows: ISSBA slightly leads Ours with an average ASR of approximately 81.80\% versus Ours' average ASR of 81.72\%, indicating their potent capabilities in compromising the models. FIBA follows closely with an average ASR of 77.88\%, demonstrating moderate effectiveness in attacks. WaNet shows a lower average ASR of about 56.88\%, suggesting lower effectiveness compared to the others. Based on these findings, ISSBA and Ours emerge as the most potent and superior attack methods, with high success rates in attacks and a substantial impact on image quality. ISSBA slightly leads due to consistently high ASR and pervasive impact across multiple models and metrics, followed closely by Ours. FIBA, on the other hand, has a lower average ASR compared to ISSBA and Ours and may also have a lower impact on image quality compared to ISSBA and Ours.

\subsection{Attack Stealthiness}
An ideal backdoor attack on face restoration models should not exacerbate the degradation of image quality. Therefore, in our study, we assess the image quality of both benign low-quality images and poison images on training sets referring to high-resolution ground truth. The evaluation results are present in Table~\ref{table:stealthy}. It is observed that these invisible attack methods, including AS-FIBA, do not significantly diminish the quality of benign images compared to inputs without attacks. This finding confirms the stealthiness of these methods, including AS-FIBA, in the context of face restoration models.

\begin{table}[h]
	\centering
 \caption{Comparison of the quality of poisoned images.}
	\begin{tabular}{c|ccc}
		\toprule
		& PSNR$_{\uparrow}$    & SSIM$_{\uparrow}$   & lPIPS$_{\downarrow}$  \\ 
		\midrule
		No Attack & 22.2082 & 0.4315 & 0.6736 \\ 
		\midrule
		WaNet     & 20.9377 & 0.4204 & 0.6398 \\ 
		ISSBA     & 22.1894 & 0.4301 & 0.6664 \\ 
		FIBA      & 21.0875 & 0.4166 & 0.6841 \\ 
		Ours   & 21.9896 & 0.4286 & 0.6754 \\ 
		\bottomrule
	\end{tabular}
	
   \label{table:stealthy}
\end{table}

\begin{figure}[h]
	\centering
	\includegraphics[width=0.45\textwidth]{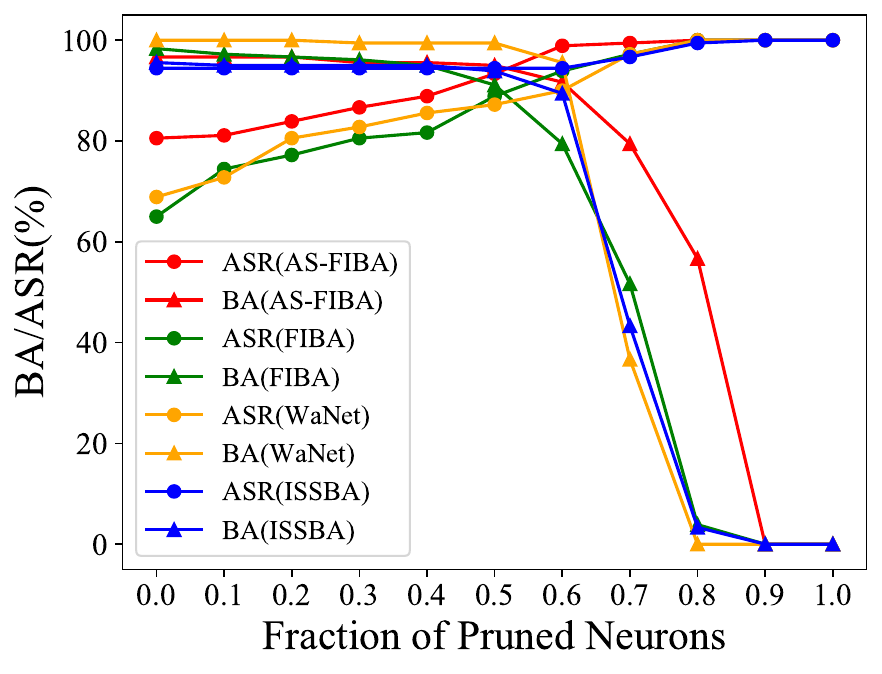}
	\caption{Benign accuracy (BA) and attack success rate (ASR) of different attack methods against pruning-based defense.}
	\label{fig:pruning_lines}
\end{figure}

\textbf{Resistance to Fine-Pruning.} We evaluate the defense of Fine-Pruning~\cite{finepruning} against various backdoor attacks on the HiFaceGAN model, focusing on their BA and ASR performances relative to neuron pruning percentages. As depicted in Fig.~\ref{fig:pruning_lines}, the BA for all attacks stand stable before pruning $60\%$ neurons. Then, they significantly decline to 0 when pruning $70\%-80\%$ neurons, where only our AS-FIBA maintains non-zero. This somehow indicates the superior robustness of our method.  However, the actual reason is that the HiFaceGAN is so sensitive to pruning that the restoration model totally failed. The rising curves of ASR also prove this view. Furthermore, we also presented a visual representation of the performance degradation of various backdoor attack methods on HiFaceGAN when faced with the Fine-Pruning defense method. Fig.~\ref{fig:pruning_imgs} shows the outputs of benign inputs and poisoned inputs at different pruning ratios. In terms of visual effects, our method produces clearer outputs compared to other methods when the pruning ratio is high, indicating that our method is less susceptible to Fine-Pruning.

\begin{figure}[h]
\centering
	\includegraphics[width=0.5\textwidth]{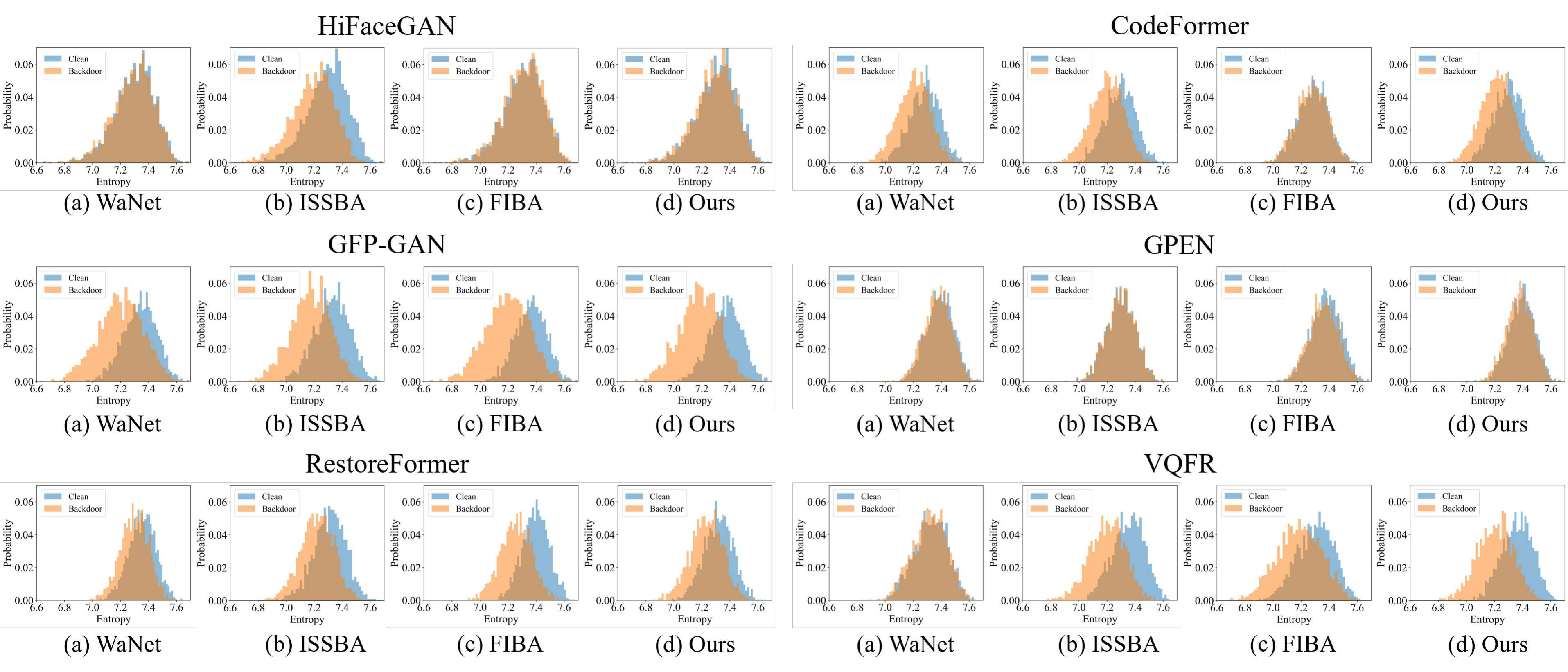}
	\caption{Entropy of STRIP defense against different attacks of six models. Best view by zooming in.}
	\label{fig:strip}
\end{figure}

\textbf{Resistance to STRIP.} STRIP~\cite{gao2019strip} introduces intentional, strong perturbations to the input and observes the output's variance. If the output remains consistent despite these perturbations, the input is likely to contain a backdoor trigger. Fig.~\ref{fig:strip} reveals the output entropy comparison of HiFaceGAN, CodeFormer, GFP-GAN, GPEN, RestoreFormer and VQFR while inputting clean and backdoor mixed images. We utilize 2,000 clean images to calculate the entropy. Taking HiFaceGAN as an illustrative instance, it is observed that only ISSBA shows a slight entropy decrease, while other backdoors maintain close to the clean inputs. This can be attributed to the degraded objective, specifically designed for the face restoration models. This kind of input-specific target results in a more subtle backdoor attack.

\subsection{Attack Behavior Analysis}
\begin{figure}[h]
\centering
	\includegraphics[width=0.45\textwidth]{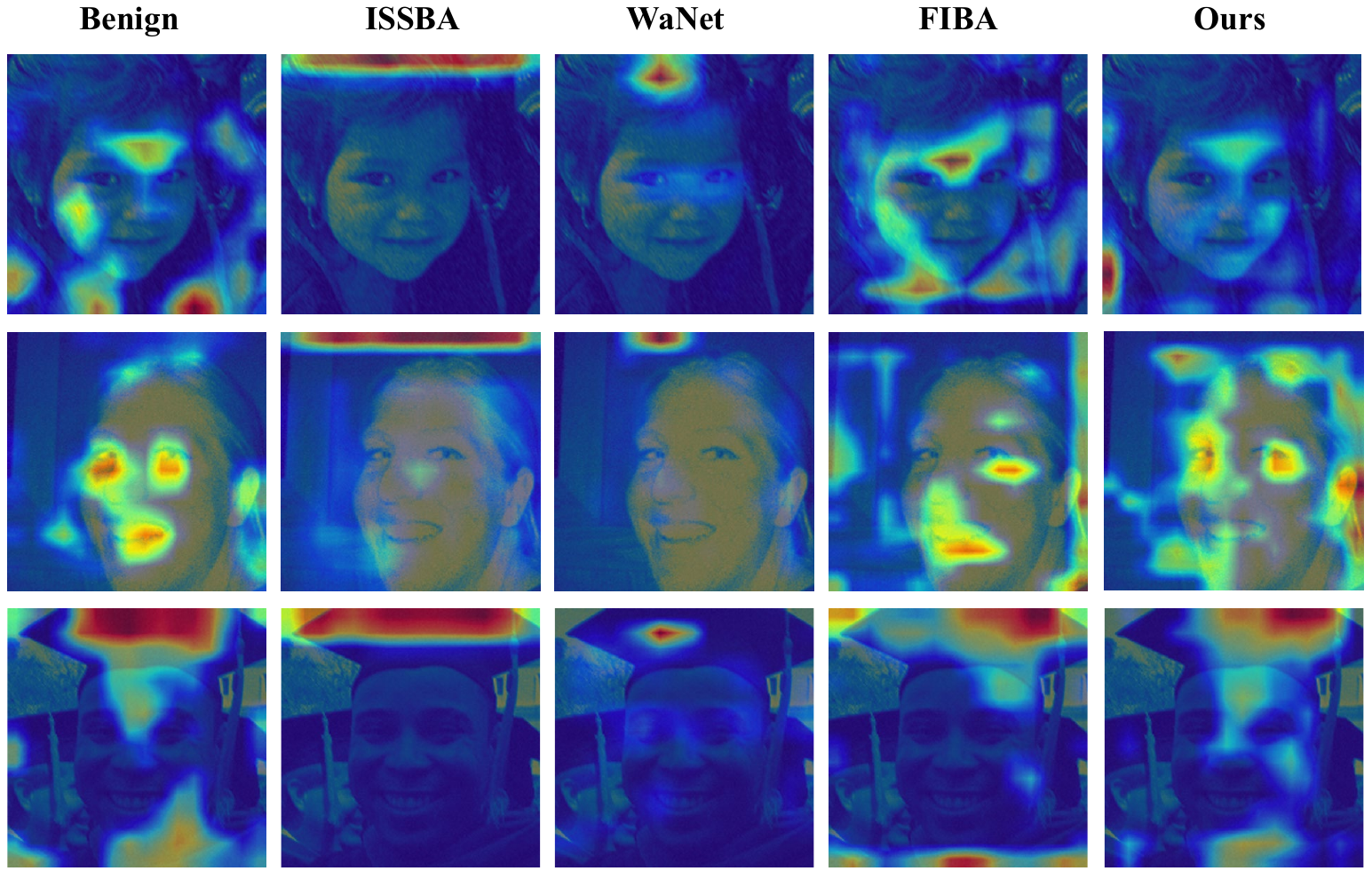}
	\caption{Visualization using Grad-CAM~\cite{selvaraju2017grad} on clean and victim models under different attacks.}
	\label{fig:grad_cam}
\end{figure}

 Grad-CAM~\cite{selvaraju2017grad} is a widely used technique that provides insights into the regions of an image that are crucial for a neural network's decision, allowing for interpretable and visually informative explanations of deep learning models. In Fig.~\ref{fig:grad_cam}, we evaluate the performance of the backdoor-injected HiFaceGAN model against these invisible attack methods. It is noticed that WaNet and ISSBA, although the visualization heat maps exhibit a certain degree of similarity between clean and backdoor images, a visible distinction still remained. However, the visualization heat maps generated by FIBA and AS-FIBA attacks display close patterns for both clean and backdoor images. While these two methods both rely on the frequency-injection, it is concluded that frequency-based backdoors have exceedingly small perturbation, resulting in minimal differences within the latent space where the heat maps are generated. This echoes the conclusion of FIBA itself. 

\subsection{Frequency-Injection Analysis}
The adaptive frequency-injection capability of AS-FIBA spurs our investigation into where the information predominantly resides: in the higher or lower frequency spectrum. Employing a technique similar to FIBA, we separate high and low frequencies in all training data and calculate the frequency distance between benign and poisoned images. As indicated in Table~\ref{table:analysis}, consistent with FIBA's design philosophy, there's a significant discrepancy in the low-frequency segment. Our AS-FIBA also has differences in the low-frequency part more than the high-frequency part. This suggests that most trigger information in AS-FIBA is also predominantly injected into the lower frequencies. Notably, the low-frequency distance in AS-FIBA is markedly less than in FIBA, \textit{i.e.} $35.5 \quad \textit{vs.} \quad 312.2$, underscoring the enhanced stealthiness of our method.

\begin{table}[h]
	\centering
 \caption{Comparison of the frequency distance (MSE) between benign and poisoned images.}
	\begin{tabular}{c|cc}
		\toprule
		& Low-Frequency & High-Frequency \\ 
		\midrule
		FIBA    & 312.20        & 0.55           \\ 
		Ours & 35.50         & 0.59           \\ 
		\bottomrule
	\end{tabular}
	
	\label{table:analysis}
\end{table}

\section{Conclusion}
In conclusion, our work has addressed the vulnerability of deep learning-based face restoration models to backdoor attacks. We introduce a novel degradation objective tailored specifically for restoration models, ensuring subtle yet impactful alterations in outputs. Our Adaptive Selective Frequency Injection Backdoor Attack (AS-FIBA) framework marks a significant leap in this domain. By generating input-specific triggers in the frequency domain, AS-FIBA seamlessly integrates these triggers into benign images, achieving a level of stealth superior to existing methods. Our extensive experiments confirm the effectiveness of our approach, positioning AS-FIBA as a critical tool for enhancing the robustness of face restoration models against sophisticated backdoor attacks.

\bibliographystyle{IEEEtran}
\bibliography{ref}

\end{document}